\documentclass{article}
\usepackage{spconf,amsmath,graphicx}
\usepackage[colorlinks=true]{hyperref}
\usepackage{multirow}
\usepackage{subcaption}
\usepackage[T1]{fontenc}
\usepackage{cite}
\usepackage{booktabs}
\usepackage{amsmath}
\usepackage{marvosym}
\usepackage{subcaption}
\usepackage{float}
\usepackage[marginal]{footmisc}

\usepackage{amssymb}
\usepackage{float}

\title{TRIM-ReID: Duplication-Aware Token Reduction and Modality-Aligned Interaction for Multi-Modal Object Re-Identification}
\name{
\begin{tabular}{c}
Wanke Xia\textsuperscript{1,*,\Letter}, 
Ruiding Zhu\textsuperscript{2,*}, 
Xingguo Xu\textsuperscript{3}, 
Zhengbo Zhang\textsuperscript{4}, \\
Dongxia Liu\textsuperscript{1},
Yuan Jin\textsuperscript{5}, 
Taojie Zhu\textsuperscript{1}, 
Yiting Zhao\textsuperscript{1}, 
Yihang Ding\textsuperscript{1}
\thanks{\textsuperscript{*}Equal Contribution.
\textsuperscript{\Letter}Correspondence: xwk25@mails.tsinghua.edu.cn;\\
Code and model weight will be released upon acceptance.
}
\end{tabular}
}
\address{
\textsuperscript{1}Tsinghua University,
\textsuperscript{2}Anhui University,
\textsuperscript{3}Dalian University of Technology,
\textsuperscript{4}CASIA,
\textsuperscript{5}SJTU
\\
}

\begin{document}
%
\maketitle
\begin{abstract}
Multi-modal object re-identification exploits complementary RGB, near-infrared (NIR), and thermal-infrared (TIR) observations to retrieve target objects. 
However, existing methods commonly employ visual encoders optimized for global image-text alignment and select tokens using learned importance scores. 
Such designs fail to preserve fine-grained identity cues or explicitly account for token redundancy, resulting in underrepresented local evidence and duplicated tokens that lead to noisy and costly cross-modal interaction.
To address this gap, we propose \textbf{TRIM-ReID}, a compact framework that unifies dense feature extraction, intra-modal token reduction, and inter-modal aligned interaction.
Specifically, semantically rich and spatially coherent patch features are extracted by \textbf{Dense Identity Representation (DIR)}, which leverages DINOv3 to preserve fine-grained identity information.
We then introduce \textbf{Token Diversity Mining (TDM)} to identify complementary local evidence and construct compact modality-specific token sets by suppressing repetitive patches while preserving informative diversity.
Retained tokens are subsequently fused by \textbf{Modal Relational Interaction (MRI)} to enable effective information exchange across modalities, while a triangular alignment loss explicitly regularizes their joint relationships to maintain cross-modal semantic consistency under independent token selection.
Extensive experiments on RGBNT201, RGBNT100, and MSVR310 demonstrate that TRIM-ReID achieves state-of-the-art performance.

\end{abstract}

\section{Introduction}

Multi-modal object re-identification (ReID) leverages complementary observations from RGB, near-infrared (NIR), and thermal-infrared (TIR) modalities to identify objects across challenging imaging conditions. Recent studies~\cite{zhang2024gait,zheng2021robust,wang2024top,zhang2024magic,wu2025lrmm,wang2025idea,zhang2025prompt,wan2026ugg,liu2026signal,xu2026stmi,li2026next} have explored selective token modeling and multi-modal fusion to improve the efficiency and discriminability of multi-modal ReID. 
However, existing methods remain limited in preserving fine-grained identity evidence while maintaining compact and semantically consistent cross-modal representations.

A key challenge lies in \textbf{how to construct informative and compact token representations}. Existing approaches often rely on visual encoders pretrained with global image-text alignment objectives~\cite{xu2026stmi,li2026next,wang2025idea}, followed by token selection based on learned importance scores~\cite{liu2026signal}. However, importance-oriented selection does not explicitly consider token redundancy, and may retain multiple highly similar patches while discarding complementary local evidence. 
Recent findings in vision-language model token pruning~\cite{wen2025stop,tan2026idpruner} further suggest that redundancy can be more critical than token importance when constructing compact representations. 
For multi-modal ReID, fine-grained identity cues are often distributed across local regions. 
Thus, token selection should not merely identify important patches, but also preserve diverse and complementary identity evidence.
Another challenge lies in \textbf{how to ensure consistent cross-modal interaction after token selection}.
Since different modalities emphasize distinct visual characteristics, independently retained tokens may exhibit heterogeneous semantic distributions. 
Direct feature fusion can facilitate information exchange, but does not necessarily guarantee semantic consistency across modalities. 
Existing methods have explored cross-modal interaction through deformable aggregation~\cite{wang2025idea}, mixture of experts~\cite{li2026next}, and hypergraph modeling~\cite{xu2026stmi}. However, the semantic consistency of independently selected modality-specific tokens remains insufficiently addressed.

To address these challenges, we propose \textbf{TRIM-ReID}, which unifies dense feature extraction, intra-modal token reduction, and inter-modal aligned interaction. \textbf{Dense Identity Representation (DIR)} leverages DINOv3 to extract spatially coherent patch features for fine-grained identity modeling. Then, \textbf{Token Diversity Mining (TDM)} constructs compact token sets by suppressing repetitive patches while preserving complementary local evidence. Finally, \textbf{Modal Relational Interaction (MRI)} uses a Transformer for cross-modal fusion, with a triangular alignment loss to enforce semantic consistency. Experiments on RGBNT201, RGBNT100, and MSVR310 validate the effectiveness of TRIM-ReID.


\begin{figure*}[t]
\centering
\includegraphics[width=1.0\textwidth]{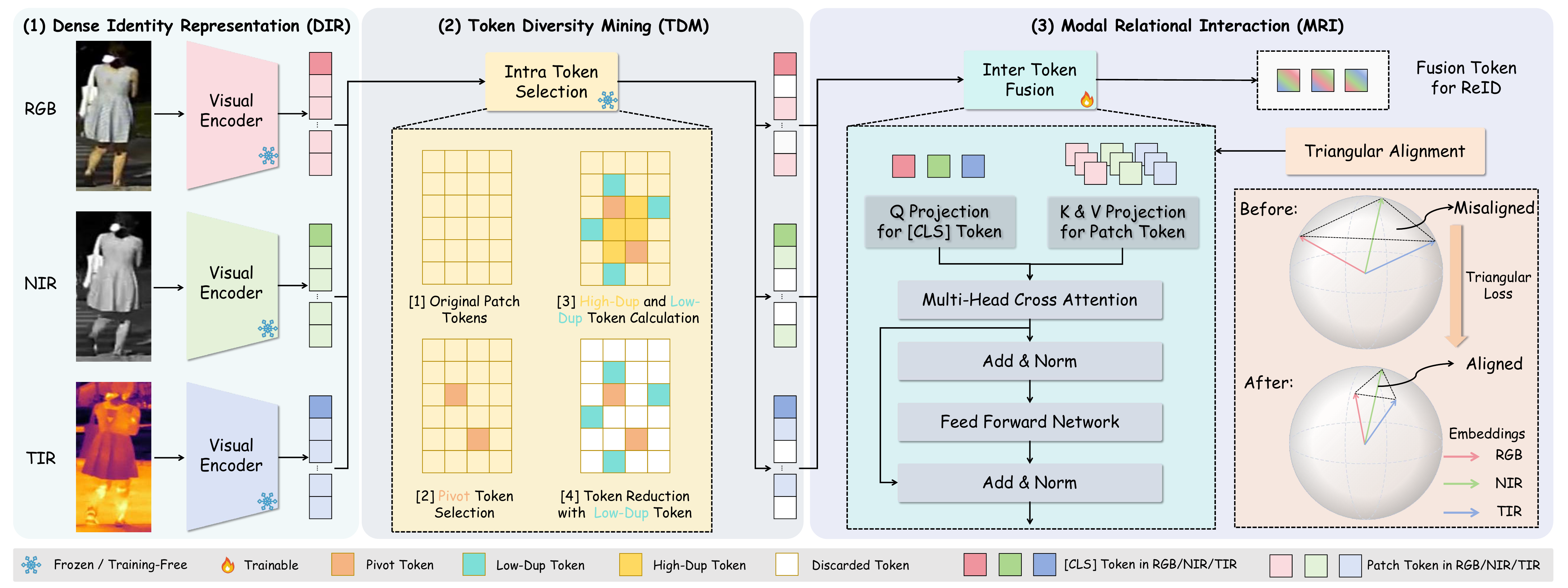}
\caption{\textbf{Overall framework of TRIM-ReID.} It comprises three core components: DIR, TDM and MRI.}
\label{fig:framework}
\end{figure*}

\section{Methodology}
\label{sec:method}

\subsection{Dense Identity Representation (DIR)}

Existing multi-modal ReID methods typically employ CLIP visual encoders pre‑trained via image‑text alignment~\cite{radford2021learning}. 
Though these models yield powerful semantic representations, their features are not explicitly optimized for dense fine‑grained identity discrimination.
Moreover, global image-text alignment can attenuate local identity cues critical for separating visually similar objects.
To obtain dense and spatially coherent visual representations, we employ DINOv3~\cite{simeoni2025dinov3} as the visual encoder. 
Given an input image $I^m$ from modality $m\in\{\mathrm{RGB},\mathrm{NIR},\mathrm{TIR}\}$, DINOv3 produces a modality-specific class token and a sequence of patch tokens:
\begin{equation}
\mathbf{f}_{cls}^{m},\,
\mathbf{F}^{m}
=
\operatorname{VisualEncoder}(I^m),
\end{equation}
where $\mathbf{f}_{cls}^{m}\in\mathbb{R}^{D}$ denotes the [CLS] token and
\begin{equation}
\mathbf{F}^{m}
=
\left[
\mathbf{f}^{m}_{1},
\mathbf{f}^{m}_{2},
\ldots,
\mathbf{f}^{m}_{N}
\right]
\in\mathbb{R}^{N\times D}
\end{equation}
denotes the corresponding patch tokens.

\subsection{Token Diversity Mining (TDM)}
Current token selection methods typically rank tokens according to their importance, which may retain highly repetitive patches while discarding complementary local evidence. 
Inspired by DART~\cite{wen2025stop}, we introduce Token Diversity Mining (TDM) to construct compact token sets according to token diversity. 
As illustrated in Fig.~\ref{fig:framework} (2), TDM is applied independently to each modality and consists of three steps: pivot token selection, duplication estimation, and token reduction.

\noindent\textbf{Pivot Token Selection.}
Given the patch tokens
$\mathbf{F}^{m}=\{\mathbf{f}_{1}^{m},\ldots,\mathbf{f}_{N}^{m}\}$,
we first randomly select a small subset of representative pivot tokens
\begin{equation}
\mathbf{P}^{m}
=
\left\{
\mathbf{p}_{1}^{m},
\ldots,
\mathbf{p}_{K}^{m}
\right\}
\subseteq
\mathbf{F}^{m},
\qquad K\ll N.
\end{equation}
The pivot tokens provide a compact reference set for estimating redundancy among the remaining patch tokens.

\noindent\textbf{Duplication Estimation.}
For each pivot token $\mathbf{p}_{k}^{m}$ and patch token $\mathbf{f}_{j}^{m}$, we measure their duplication by cosine similarity:
\begin{equation}
d_{k,j}^{m}
=
\frac{
(\mathbf{p}_{k}^{m})^{\top}\mathbf{f}_{j}^{m}
}{
\|\mathbf{p}_{k}^{m}\|_2
\|\mathbf{f}_{j}^{m}\|_2
}.
\end{equation}
Two tokens are considered $\epsilon$-duplicates when
$d_{k,j}^{m}>\epsilon_{k}^{m}$, where $\epsilon_{k}^{m}$ is dynamically determined for each pivot token according to the desired token reduction ratio. 
For each pivot $\mathbf{p}_{k}^{m}$, we therefore construct the corresponding retained set as
\begin{equation}
\mathbf{R}_{k}^{m}
=
\left\{
\mathbf{f}_{j}^{m}
\mid
d_{k,j}^{m}\leq\epsilon_{k}^{m}
\right\}.
\end{equation}
A higher duplication score indicates stronger redundancy with the pivot, whereas a lower score suggests more complementary visual information. 
Duplication estimation requires no additional training due to computation of frozen features.

\noindent\textbf{Token Reduction.}
The final retained token set for modality $m$ is obtained by
\begin{equation}
\mathbf{R}^{m}
=
\mathbf{P}^{m}
\cup
\left(
\bigcup_{k=1}^{K}
\mathbf{R}_{k}^{m}
\right),
\end{equation}
where $\mathbf{P}^{m}$ denotes the set of pivot tokens. 
This procedure removes highly duplicated patches (i.e., High-Dup Tokens) while preserving tokens that provide complementary local evidence (i.e., Low-Dup Tokens).




\begin{figure}[t]
\centering
\includegraphics[width=\columnwidth]{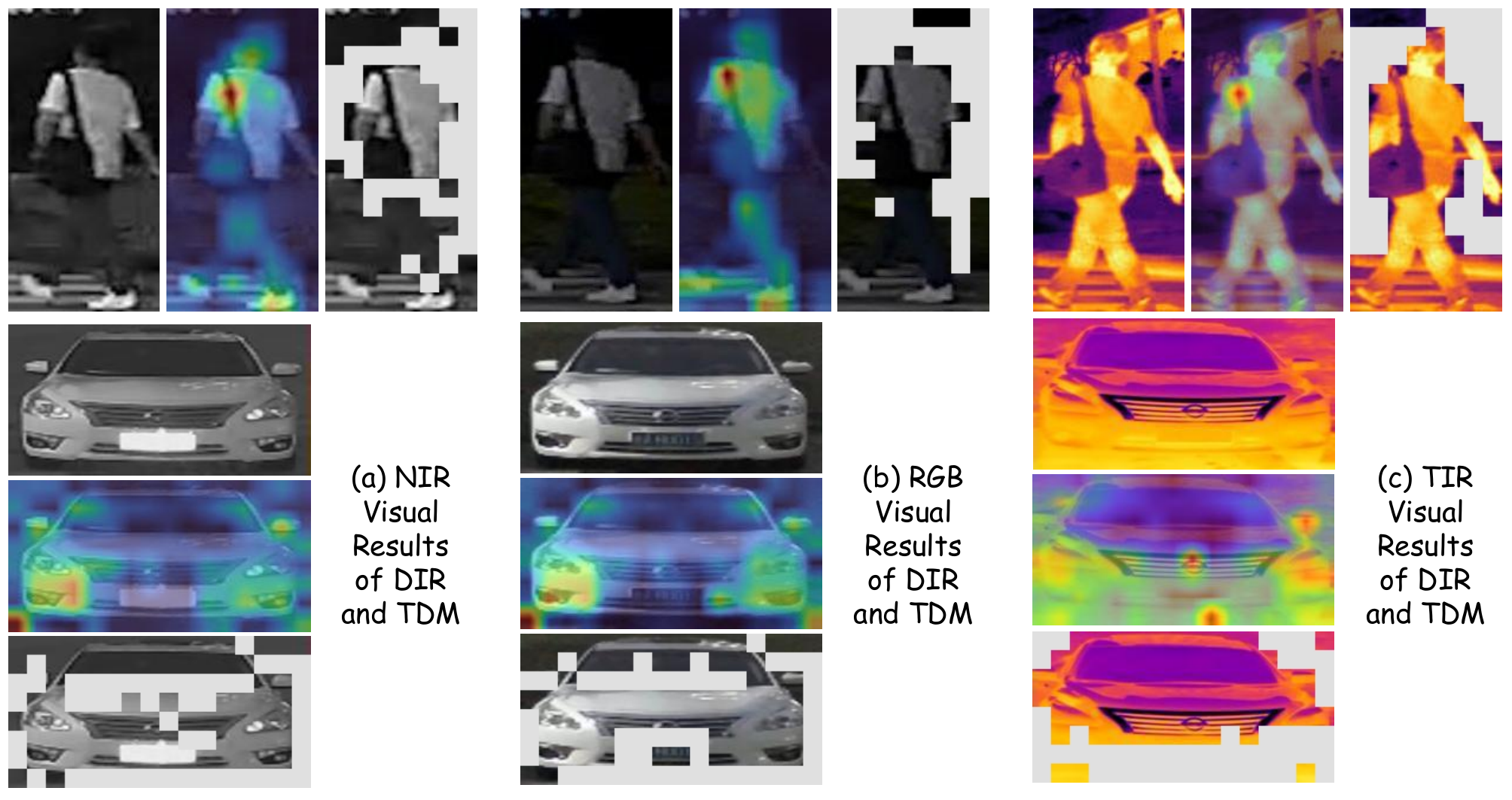}
\caption{Visualization of attention maps and token reduction.}
\label{fig:vis}
\end{figure}

\begin{table*}[t]
\centering
\caption{Performance comparison of various methods on the datasets of RGBNT201, RGBNT100, and MSVR310. The best
and second results are in \textbf{bold} and \underline{underlined}, respectively. Higher values indicate better performance for all metrics.}
\label{tab:main_results}
\resizebox{0.75\textwidth}{!}{
\begin{tabular}{ll cccc cc cc}
\toprule
\multirow{2}{*}{Methods} & \multirow{2}{*}{Venue}
& \multicolumn{4}{c}{RGBNT201}
& \multicolumn{2}{c}{RGBNT100}
& \multicolumn{2}{c}{MSVR310} \\
\cmidrule(lr){3-6} \cmidrule(lr){7-8} \cmidrule(lr){9-10}
& & mAP & R-1 & R-5 & R-10 & mAP & R-1 & mAP & R-1 \\
\midrule
PFNet~\cite{zheng2021robust} & AAAI'21 & 38.5 & 38.9 & 52.0 & 58.4 & 68.1 & 94.1 & 23.5 & 37.4 \\
TOP-ReID~\cite{wang2024top} & AAAI'24 & 72.3 & 76.6 & 84.7 & 89.4 & 81.2 & 96.4 & 35.9 & 44.6 \\
EDITOR~\cite{zhang2024magic} & CVPR'24 & 66.5 & 68.3 & 81.1 & 88.2 & 82.1 & 96.4 & 39.0 & 49.3 \\
LRMM~\cite{wu2025lrmm} & ESWA'25 & 52.3 & 53.4 & 64.6 & 73.2 & 78.6 & 96.7 & 36.7 & 49.7 \\
IDEA~\cite{wang2025idea} & CVPR'25
& 80.2 & 82.1 & 90.0 & 93.3 & 87.2 & 96.5 & 47.0 & 62.4 \\
PromptMA~\cite{zhang2025prompt} & TIP'25 & 78.4 & 80.9 & 87.0 & 88.9 & 85.3 & 97.4 & 55.2 & 64.5 \\
UGG-ReID~\cite{wan2026ugg} & NeurIPS'25 & 81.2 & \underline{86.8} & \underline{92.0} & \underline{94.7} & 88.0 & \underline{98.1} & 60.1 & 78.0 \\
Signal~\cite{liu2026signal} & AAAI'26
& 80.3 & 85.2 & 91.4 & 93.7 & 86.3 & 97.6 & 53.6 & 71.9 \\
STMI~\cite{xu2026stmi} & AAAI'26 & 81.2 & 83.4 & 90.2 & 91.6 & \underline{89.1} & 97.1 & \underline{64.8} & 76.1 \\
NEXT~\cite{li2026next} & TIFS'26 & \underline{82.4} & 86.6 & \underline{92.0} & \underline{94.7} & 88.2 & 97.7 & 60.8 & \underline{79.0} \\
TRIM-ReID (Ours) & - & \textbf{85.2} & \textbf{88.0}  & \textbf{92.7}  & \textbf{95.5}  &  \textbf{89.2} & \textbf{98.7}  & \textbf{68.8} &  \textbf{83.1} \\
\bottomrule
\end{tabular}
}
\end{table*}

\subsection{Modal Relational Interaction (MRI)}
\label{sec:mri}

\noindent\textbf{Inter Token Fusion.}
Each modality retains a compact set of informative patch tokens and its corresponding [CLS] token after TDM. 
We independently project each [CLS] token into a query embedding and project the retained patch tokens from each modality into keys and values. 
Each modality-specific query then attends to the patch tokens from all three modalities.
Specifically, for each modality 
$m\in\{\mathrm{RGB},\mathrm{NIR},\mathrm{TIR}\}$,
let $\mathbf{f}_{cls}^{m}\in\mathbb{R}^{D}$ denote the [CLS] token and
$\mathbf{R}^{m}\in\mathbb{R}^{N_m\times D}$ denote the retained patch tokens after TDM. 
For each attention head $h\in\{1,\ldots,H\}$, the query, key, and value are as:
\begin{equation}
\mathbf{Q}_{h}^{m}
=
\mathbf{f}_{cls}^{m}\mathbf{W}_{Q}^{h},
\;
\mathbf{K}_{h}^{m}
=
\mathbf{R}^{m}\mathbf{W}_{K}^{h},
\;
\mathbf{V}_{h}^{m}
=
\mathbf{R}^{m}\mathbf{W}_{V}^{h},
\end{equation}
where $\mathbf{W}_{Q}^{h}$, $\mathbf{W}_{K}^{h}$, and $\mathbf{W}_{V}^{h}$ are learnable projection matrices for head $h$.
To enable cross-modal interaction, the key and value representations from all modalities are concatenated:
\begin{equation}
\mathbf{K}_{h}
=
[
\mathbf{K}_{h}^{rgb};
\mathbf{K}_{h}^{nir};
\mathbf{K}_{h}^{tir}
],
\;
\mathbf{V}_{h}
=
[
\mathbf{V}_{h}^{rgb};
\mathbf{V}_{h}^{nir};
\mathbf{V}_{h}^{tir}
].
\end{equation}
The attention output for modality $m$ and head $h$ is given by:
\begin{equation}
\mathbf{A}_{h}^{m}
=
\operatorname{Softmax}
\left(
\frac{
\mathbf{Q}_{h}^{m}
\mathbf{K}_{h}^{\top}
}{
\sqrt{d_h}
}
\right)
\mathbf{V}_{h},
\end{equation}
where $d_h$ denotes the dimension of each attention head. 
The outputs of all heads are concatenated and projected as:
\begin{equation}
\mathbf{h}^{m}
=
\operatorname{Concat}
\left(
\mathbf{A}_{1}^{m},
\ldots,
\mathbf{A}_{H}^{m}
\right)
\mathbf{W}_{O}.
\end{equation}
We then apply a residual connection and layer normalization:
\begin{equation}
\mathbf{z}^{m}
=
\operatorname{LayerNorm}
\left(
\mathbf{f}_{cls}^{m}
+
\mathbf{h}^{m}
\right),
\end{equation}
and $\mathbf{z}^{m}$ is further refined by a feed forward network:
\begin{equation}
\mathbf{f}_{fusion}^{m}
=
\operatorname{LayerNorm}
\left(
\mathbf{z}^{m}
+
\operatorname{FFN}(\mathbf{z}^{m})
\right).
\end{equation}
Finally $\mathbf{f}_{fusion}^{m}$ serves as the modality-aware fusion feature.

\noindent \textbf{Triangular Alignment.}
We further introduce a triangular alignment that jointly aligns the three modalities in their shared embedding space based on TRIANGLE~\cite{cicchetti2026triangle}.
We first normalize them to the unit hypersphere and regard the three embeddings as vertices in the shared $D$-dimensional space. The three vertices form a triangle, whose area provides a direct measure of their joint alignment. 
We use each modality as the anchor in turn and include both matching directions.
Taking the RGB anchor for example, let
\begin{equation}
\mathbf{u}=\mathbf{z}^{rgb}-\mathbf{z}^{nir},\qquad
\mathbf{v}=\mathbf{z}^{rgb}-\mathbf{z}^{tir},
\end{equation}
where $\mathbf{z}^{rgb}$, $\mathbf{z}^{nir}$, and $\mathbf{z}^{tir}\in\mathbb{R}^{D}$ denote the modality-specific representations obtained from the retained tokens.
The corresponding triangular area is then defined as
\begin{equation}
\mathcal{A}_{rgb}
=
\frac{1}{2}
\sqrt{[
\langle\mathbf{u},\mathbf{u}\rangle
\langle\mathbf{v},\mathbf{v}\rangle
-
\langle\mathbf{u},\mathbf{v}\rangle^2
]_{+}},
\end{equation}
where $[x]_{+}=\max(x,0)$.
A smaller area indicates that the three modality representations are geometrically closer and thus better aligned, while a larger area implies greater cross-modal discrepancy. Finally, the alignment loss is defined as:
\begin{equation}
\mathcal{L}_{align}
=Mean(\mathcal{A}_{rgb}+\mathcal{A}_{nir}+\mathcal{A}_{tir}).
\end{equation}

\subsection{Training Objectives}
The overall objective combines identity and alignment losses:
\begin{equation}
\mathcal{L}
=
\mathcal{L}_{\mathrm{CE}}
+
\mathcal{L}_{\mathrm{triplet}}
+
\lambda\mathcal{L}_{\mathrm{align}},
\end{equation}
where $\mathcal{L}_{\mathrm{CE}}$, $\mathcal{L}_{\mathrm{triplet}}$, and $\mathcal{L}_{\mathrm{align}}$ denote the label-smoothed cross-entropy, triplet, and triangular alignment losses, respectively, and $\lambda$ controls the alignment weight.

\begin{table}[t]
\centering
\caption{Ablation on the proposed modules of TRIM-ReID.}
\label{tab:ablation_module}
\begin{tabular}{l ccc cc}
\toprule
\multirow{2}{*}{Variants} & \multicolumn{3}{c}{Modules} & \multicolumn{2}{c}{Metrics} \\
\cmidrule(lr){2-4} \cmidrule(lr){5-6}
& DIR & TDM & MRI & mAP & R-1 \\
\midrule
A & $\checkmark$ & $\times$ & $\times$ & 76.4 & 79.1 \\
B & $\checkmark$ & $\checkmark$ & $\times$ & 80.6 & 84.6 \\
C & $\checkmark$ & $\checkmark$ & $\checkmark$ & 85.2 & 88.0 \\
\bottomrule
\end{tabular}
\end{table}

\begin{table}[t]
\centering
\caption{Comparison of different visual encoders.}
\label{tab:ablation_encoder}
\begin{tabular}{l cccc}
\toprule
Encoders & mAP & R-1 & R-5 & R-10 \\
\midrule
CLIP~\cite{radford2021learning} & 74.8 & 80.5 & 87.2 & 90.8 \\
DINOv3~\cite{simeoni2025dinov3} & 85.2 & 88.0  & 92.7  & 95.5 \\
\bottomrule
\end{tabular}
\end{table}

\section{Experiments}
\label{sec:experiment}

\subsection{Experiment Setup}
\noindent \textbf{Implementation Details.}
Input images are resized to $256\times128$ for RGBNT201 and $128\times256$ for RGBNT100 and MSVR310, with random flipping, cropping, and erasing applied for data augmentation. 
The batch sizes are set to 64, 64, and 128 for RGBNT201, MSVR310, and RGBNT100, respectively, with 8, 4, and 16 images sampled per identity. 
We implement TRIM-ReID in PyTorch using two NVIDIA RTX 3090 GPUs and train it with Adam for 50 epochs. 
The learning rate is set to $5\times10^{-4}$ for RGBNT201 and RGBNT100, and $5\times10^{-5}$ for MSVR310. 
For TDM, we use 4 pivot tokens for RGBNT201 and RGBNT100 and 8 for MSVR310, while the number of retained tokens is fixed to 80. 
$\lambda$ is set to 0.04.

\noindent \textbf{Evaluation Metrics.}
We apply the Mean Average Precision (mAP)
and Rank-K (K = 1, 5, 10) matching accuracy as evaluation
metrics to align with previous studies.

\subsection{Results and Analysis}
\noindent \textbf{Main Results.}
Table~\ref{tab:main_results} shows that TRIM-ReID consistently achieves the best performance across all three datasets. 
On RGBNT201, it reaches 85.2\% mAP and 88.0\% R-1, gaining 2.8 and 1.2 points compared to NEXT and UGG-ReID, respectively.
On RGBNT100, it obtains 89.2\% mAP and 98.7\% R-1, surpassing by 0.1 and 0.6 points compared to STMI and UGG-ReID, respectively.
Notably, on MSVR310, it achieves 68.8\% mAP and 83.1\% R-1, exceeding by 4.0 and 4.1 points compared to STMI and NEXT, respectively.
These results indicate that duplication-aware token reduction and modality-aligned interaction are particularly beneficial in challenging scenarios of person and vehicle re-identification.

\noindent \textbf{Visualization Results.}
We provide visualizations of the attention maps from DIR and the token reduction results of TDM (Fig.~\ref{fig:vis}), illustrating their respective abilities to capture dense representations and preserve informative visual tokens.

\subsection{Ablation Studies}
We conduct all ablation studies on the RGBNT201 dataset.

\noindent \textbf{Effects of Proposed Modules.}
Module-wise ablation in Table~\ref{tab:ablation_module} validates each component. The DIR-only baseline yields 76.4\% mAP and 79.1\% Rank-1. Adding TDM improves performance to 80.6\% mAP and 84.6\% Rank-1, while full MRI further boosts results to 85.2\% mAP and 88.0\% Rank-1, confirming TDM’s compact diverse tokens enable effective cross‑modal interaction and the alignment loss preserves cross‑modal semantic consistency.

\noindent \textbf{Effects of Different Visual Encoders.}
Table~\ref{tab:ablation_encoder} compares different visual encoders. DINOv3 achieves the best performance, indicating its effectiveness in visual features.

\noindent \textbf{Effects of the Number of Retained Tokens and Pivot Tokens in TDM.}
Table~\ref{tab:ablation_k} and Table~\ref{tab:ablation_pivot} discuss the effects of the number of retained tokens ($N_r$) and pivot tokens ($N_p$) in TDM. 
The best mAP is achieved when $N_r=80$ and $N_p=4$ for these two ablations, respectively.

\begin{table}[t]
\centering
\caption{Ablation on the number of retained tokens in TDM.}
\label{tab:ablation_k}
\begin{tabular}{l cccc}
\toprule
Number Setting & mAP & R-1 & R-5 & R-10 \\
\midrule
$N_r = 32$ & 81.5 & 84.8 & 89.1 & 91.9 \\
$N_r = 48$ & 82.6 & \textbf{89.0} & 93.0 & 95.0 \\
$N_r = 64$ & 80.3 & 83.4 & 89.6 & 92.7 \\
$N_r = 80$ & \textbf{85.2} & 88.0  & 92.7  & \textbf{95.5}\\
$N_r = 96$ & 80.7 & 83.9 & 91.3 & 94.1 \\
$N_r = 112$ & 83.3 & 87.0 & 91.5 & 94.0 \\
$N_r = 128$ & 84.0 & 88.5 & \textbf{93.5} & 95.0 \\
\bottomrule
\end{tabular}
\end{table}


\begin{table}[t]
\centering
\caption{Ablation on the number of pivot tokens in TDM.}
\label{tab:ablation_pivot}
\begin{tabular}{l cccc}
\toprule
Number Setting & mAP & R-1 & R-5 & R-10 \\
\midrule
$N_p = 2$ & 82.1 & 82.8 & 90.7 & 94.4 \\
$N_p = 4$ & \textbf{85.2} & \textbf{88.0}  & \textbf{92.7}  & \textbf{95.5}\\
$N_p = 8$ & 83.7 & 86.4 & \textbf{92.7} & 93.9 \\
\bottomrule
\end{tabular}
\end{table}

\section{Conclusion}
\label{sec:conclusion}
We present \textbf{TRIM-ReID}, a compact framework for multi-modal object ReID that addresses redundant token selection and cross-modal inconsistency. By combining DIR, TDM and MRI, TRIM-ReID preserves fine-grained identity evidence, reduces redundant tokens, and promotes consistent cross-modal interaction. Extensive experiments on 3 datasets demonstrate its state-of-the-art performance.


\section{Acknowledgement}
This work was partly supported by the Special Foundations for the Development of Strategic Emerging Industries of Shenzhen (No.KJZD20231023094700001) and the Shenzhen-Tsinghua Special Project for Fundamental \& Frontier Research in Artificial Intelligence (No.AI2026018).
This work also sincerely thanks Wenming Yang from Tsinghua University for his invaluable support.

\bibliographystyle{IEEEbib}
\bibliography{refs}

@inproceedings{liu2026signal,
  title={Signal: Selective interaction and global-local alignment for multi-modal object re-identification},
  author={Liu, Yangyang and Wang, Yuhao and Zhang, Pingping},
  booktitle={Proceedings of the AAAI Conference on Artificial Intelligence},
  volume={40},
  number={9},
  pages={7359--7367},
  year={2026}
}

@article{tan2026idpruner,
  title={IDPruner: Harmonizing Importance and Diversity in Visual Token Pruning for MLLMs},
  author={Tan, Yifan and Sun, Yifu and Huang, Shirui and Liu, Hong and Yu, Guanghua and Zhu, Jianchen and Deng, Yangdong},
  journal={arXiv preprint arXiv:2602.13315},
  year={2026}
}

@article{cicchetti2026triangle,
  title={A triangle enables multimodal alignment beyond cosine similarity},
  author={Cicchetti, Giordano and Grassucci, Eleonora and Comminiello, Danilo},
  journal={Advances in Neural Information Processing Systems},
  volume={38},
  pages={112004--112022},
  year={2025}
}

@inproceedings{radford2021learning,
  title={Learning transferable visual models from natural language supervision},
  author={Radford, Alec and Kim, Jong Wook and Hallacy, Chris and Ramesh, Aditya and Goh, Gabriel and Agarwal, Sandhini and Sastry, Girish and Askell, Amanda and Mishkin, Pamela and Clark, Jack and others},
  booktitle={International conference on machine learning},
  pages={8748--8763},
  year={2021},
  organization={PmLR}
}

@article{simeoni2025dinov3,
  title={Dinov3},
  author={Sim{\'e}oni, Oriane and Vo, Huy V and Seitzer, Maximilian and Baldassarre, Federico and Oquab, Maxime and Jose, Cijo and Khalidov, Vasil and Szafraniec, Marc and Yi, Seungeun and Ramamonjisoa, Micha{\"e}l and others},
  journal={arXiv preprint arXiv:2508.10104},
  year={2025}
}

@inproceedings{wen2025stop,
  title={Stop looking for “important tokens” in multimodal language models: Duplication matters more},
  author={Wen, Zichen and Gao, Yifeng and Wang, Shaobo and Zhang, Junyuan and Zhang, Qintong and Li, Weijia and He, Conghui and Zhang, Linfeng},
  booktitle={Proceedings of the 2025 Conference on Empirical Methods in Natural Language Processing},
  pages={9961--9980},
  year={2025}
}

@article{zhang2024gait,
  title={Gait recognition for farm workers via multi-scale temporal feature perception},
  author={Zhang, Zhengbo and Xiao, Yaxu and Chen, Zizhi and Chen, Xin and Li, Xiang},
  journal={Computers and Electronics in Agriculture},
  volume={226},
  note={art. no. 109353},
  year={2024},
  publisher={Elsevier}
}

@article{wu2025lrmm,
  title={LRMM: Low rank multi-scale multi-modal fusion for person re-identification based on RGB-NI-TI},
  author={Wu, Di and Liu, Zhihui and Chen, Zihan and Gan, Shenglong and Tan, Kaiwen and Wan, Qin and Wang, Yaonan},
  journal={Expert Systems with Applications},
  volume={263},
  note={art. no. 125716},
  year={2025},
  publisher={Elsevier}
}

@inproceedings{zheng2021robust,
  title={Robust multi-modality person re-identification},
  author={Zheng, Aihua and Wang, Zi and Chen, Zihan and Li, Chenglong and Tang, Jin},
  booktitle={Proceedings of the AAAI conference on artificial intelligence},
  volume={35},
  number={4},
  pages={3529--3537},
  year={2021}
}

@inproceedings{zhang2024magic,
  title={Magic tokens: Select diverse tokens for multi-modal object re-identification},
  author={Zhang, Pingping and Wang, Yuhao and Liu, Yang and Tu, Zhengzheng and Lu, Huchuan},
  booktitle={2024 IEEE/CVF Conference on Computer Vision and Pattern Recognition (CVPR)},
  pages={17117--17126},
  year={2024},
  organization={IEEE}
}

@inproceedings{wang2024top,
  title={Top-reid: Multi-spectral object re-identification with token permutation},
  author={Wang, Yuhao and Liu, Xuehu and Zhang, Pingping and Lu, Hu and Tu, Zhengzheng and Lu, Huchuan},
  booktitle={Proceedings of the AAAI conference on artificial intelligence},
  volume={38},
  number={6},
  pages={5758--5766},
  year={2024}
}

@article{li2026next,
  title={NEXT: Multi-grained mixture of experts via text-modulation for multi-modal object re-identification},
  author={Li, Shihao and Huang, Huaibo and Duan, Junxian and Zheng, Aihua and Tang, Jin and Ma, Jixin},
  journal={IEEE Transactions on Information Forensics and Security},
  year={2026},
  publisher={IEEE}
}

@article{wan2026ugg,
  title={Ugg-reid: Uncertainty-guided graph model for multi-modal object re-identification},
  author={Wan, Xixi and Zheng, Aihua and Jiang, Bo and Wang, Beibei and Li, Chenglong and Tang, Jin},
  journal={Advances in Neural Information Processing Systems},
  volume={38},
  pages={80232--80255},
  year={2025}
}

@inproceedings{xu2026stmi,
  title={STMI: Segmentation-Guided Token Modulation with Cross-Modal Hypergraph Interaction for Multi-Modal Object Re-Identification},
  author={Xu, Xingguo and Liu, Zhanyu and Zhou, Weixiang and Gao, Yuansheng and Cao, Junjie and Wang, Yuhao and Luo, Jixiang and Zhang, Dell},
  booktitle={Proceedings of the AAAI Conference on Artificial Intelligence},
  volume={40},
  number={14},
  pages={11433--11441},
  year={2026}
}

@article{zhang2025prompt,
  title={Prompt-based modality alignment for effective multi-modal object re-identification},
  author={Zhang, Shizhou and Luo, Wenlong and Cheng, De and Xing, Yinghui and Liang, Guoqiang and Wang, Peng and Zhang, Yanning},
  journal={IEEE Transactions on Image Processing},
  volume={34},
  pages={2450--2462},
  year={2025},
  publisher={IEEE}
}

@inproceedings{wang2025idea,
  title={Idea: Inverted text with cooperative deformable aggregation for multi-modal object re-identification},
  author={Wang, Yuhao and Lv, Yongfeng and Zhang, Pingping and Lu, Huchuan},
  booktitle={2025 IEEE/CVF Conference on Computer Vision and Pattern Recognition (CVPR)},
  pages={29701--29710},
  year={2025},
  organization={IEEE}
}

\end{document}